\documentclass[11pt]{article}
\usepackage{a4}
\usepackage{colacl}
\usepackage{xspace}
\usepackage{amsfonts} 

\newcommand{\negskip}{\vspace{-\smallskipamount}}
\newcommand{\TS}{T\&S\xspace}

\newcommand{\defeq}{\stackrel{\rm def}{=}}
\newcommand{\angles}[1]{\langle #1 \rangle}

\newcommand{\Gen}{\mbox{\sf Gen}\xspace}
\newcommand{\Con}{\mbox{$\cal C$}\xspace}
\newcommand{\Rul}{\mbox{$\cal R$}\xspace}
\newcommand{\Gram}{\mbox{$\cal G$}\xspace}
\newcommand{\Lex}{\mbox{$\cal L$}\xspace}

\newcommand{\Off}{\mbox{\sc Off}}
\newcommand{\Early}{\mbox{\sc Early}}
\newcommand{\Project}{\mbox{\sc Project}}
\newcommand{\Short}{\mbox{\sc Short}}
\newcommand{\Move}{\mbox{\sc Move}}
\newcommand{\Accept}{\mbox{\sc Accept}}

\newcommand{\Opt}{\mbox{\sc Opt}\xspace}
\newcommand{\OptVal}{\mbox{\sc OptVal}\xspace}
\newcommand{\OptValZ}{\mbox{\sc OptValZ}\xspace}
\newcommand{\Rcd}{\mbox{\sc Rcd}\xspace}
\newcommand{\RcdAll}{\mbox{\sc RcdAll}\xspace}
\newcommand{\Beatable}{\mbox{\sc Beatable}\xspace}
\newcommand{\Best}{\mbox{\sc Best}\xspace}
\newcommand{\Check}{\mbox{\sc Check}\xspace}
\newcommand{\CheckSSet}{\mbox{\sc CheckSSet}\xspace}
\newcommand{\Rankable}{\mbox{\sc Rankable}\xspace}
\newcommand{\RankableSSet}{\mbox{\sc RankableSSet}\xspace}
\newcommand{\TspVal}{\mbox{\sc TspVal}\xspace}
\newcommand{\Msa}{\mbox{\sc Msa}\xspace}
\newcommand{\Msalsb}{\mbox{{\sc Msa}$_{lsb}$}\xspace}
\newcommand{\CNFSat}{\mbox{\sc CNF-Sat}\xspace}
\newcommand{\Sat}{\mbox{\sc Sat}\xspace}
\newcommand{\SatUnsat}{\mbox{\sc Sat-Unsat}\xspace}
\newcommand{\Range}{\mbox{\sc Range}\xspace}
\newcommand{\QSatTwo}{\mbox{\sc QSat$_2$}\xspace}
\newcommand{\OrderableSSet}{\mbox{\sc OrderableSSet}\xspace}
\newcommand{\Orderable}{\mbox{\sc Orderable}\xspace}

\newcommand{\OptP}{\mbox{\sf OptP}\xspace}
\newcommand{\Pol}{\mbox{\sf P}\xspace}
\newcommand{\NP}{\mbox{\sf NP}\xspace}

\newcommand{\coNP}{\mbox{\sf coNP}\xspace}
\newcommand{\FP}{\mbox{\sf FP}\xspace}

\begin{document}
\thispagestyle{plain}
\setcounter{page}{22}

\title{\raisebox{0.75\baselineskip}[0in][0in]{\parbox{\linewidth}{\scriptsize\tt\begin{flushright}
        {\bf In:} Eisner, J., L. Karttunen and A.
        Th\'{e}riault (eds.), {\em Finite-State Phonology:\@
          Proc.\@ of the 5th Workshop \\ of the ACL Special
          Interest Group in Computational Phonology (SIGPHON)}, pp.\@
        22-33, Luxembourg, Aug.\@ 2000.
       [Online proceedings version:\ small corrections and clarifications to printed version.] 
       \end{flushright}}} \\
       Easy and Hard Constraint Ranking in Optimality Theory:%
       \thanks{Many thanks go to
               Lane and Edith Hemaspaandra for references to the complexity
               literature, and to Bruce Tesar for comments on an earlier draft.}
       \\ \large Algorithms and Complexity}
\author{Jason Eisner \\ 
        Dept. of Computer Science / University of Rochester \\
        Rochester, NY  14607-0226 USA / {\tt jason@cs.rochester.edu}}

\maketitle
\begin{abstract}
\small 
We consider the problem of ranking a set of OT constraints in a manner
consistent with data. (1) We speed up Tesar and Smolensky's RCD
algorithm to be linear on the number of constraints.  This finds a
ranking so each attested form $x_i$ beats or ties a particular
competitor $y_i$.  (2) We also generalize RCD so each $x_i$ beats or
ties {\em all} possible competitors.

Alas, neither ranking as in (2) nor even generation has any
polynomial algorithm unless $\Pol=\NP$---i.e., one cannot improve
qualitatively upon brute force:
(3) Merely checking that a {\em single} (given) ranking is consistent
with given forms is \coNP-complete if the surface forms are fully
observed and $\Delta_2^p$-complete if not.  Indeed, OT generation is
\OptP-complete.  (4) As for ranking, determining whether {\em any}
consistent ranking exists is $\coNP$-hard (but in $\Delta_2^p$) if the
forms are fully observed, and $\Sigma_2^p$-complete if not.

Finally, we show (5) generation and ranking are easier in derivational
theories: \Pol, and \NP-complete.

\end{abstract}

\section{Introduction}\label{sec:intro}

Optimality Theory (OT) is a grammatical paradigm that was introduced
by Prince and Smolensky
\shortcite{prince-smolensky-1993} and suggests various computational
questions, including learnability.

Following Gold \shortcite{gold-1967} we might ask: Is the language class
$\{L(\Gram): \Gram$ is an OT grammar$\}$ learnable in the limit?  That
is, is there a learning algorithm that will converge on any
OT-describable language $L(\Gram)$ if presented with an enumeration of
its grammatical forms?

In this paper we consider an orthogonal question that has been
extensively investigated by Tesar and Smolensky
\shortcite{tesar-smolensky-1996-long}, henceforth \TS.  Rather than
asking whether a learner can eventually find an OT grammar
compatible with an unbounded set of positive data, we ask: {\em How
efficiently} can it find a grammar (if one exists) compatible with a
{\em finite} set of positive data?

Sections~3--5 present successively more realistic versions of the
problem (sketched in the abstract).  The easiest version turns out to
be easier than previously known.  The harder versions turn out to be
harder than previously known.

\section{Formalism}\label{sec:formalism}

An OT grammar \Gram consists of three elements, any or all
of which may need to be learned:
\begin{itemize}
\item a set \Lex of {\bf underlying forms} produced by a lexicon or morphology,
\item a function \Gen that maps any underlying form to a set of
      {\bf candidates}, and
\item a vector $\vec{C} = \angles{C_1, C_2, \ldots C_n}$ of {\bf constraints},
      each of which is a function from candidates to the natural
      numbers $\mathbb{N}$.  
\end{itemize}

$C_i$ is said to {\bf rank} higher than (or {\bf outrank}) $C_j$ in
$\vec{C}$ iff $i < j$.  We say $x$ {\bf satisfies} $C_i$ if
$C_i(x)=0$, else $x$ {\bf violates} $C_i$.

The grammar \Gram defines a relation that maps each $u \in \Lex$ to
the candidate(s) $x \in \Gen(u)$ for which the vector $\vec{C}(x)
\defeq \angles{C_1(x), C_2(x), \ldots C_n(x)}$ is lexicographically
minimal.  Such candidates are called {\bf optimal}.

One might then say that the grammatical forms are the pairs $(u,x)$ of
this relation.  But for simplicity of notation and without loss of
generality, we will suppose that the candidates $x$ are rich enough
that $u$ can always be recovered from $x$.\footnote{This is necessary
  in any case if $C_j(x)$ is to depend on (all of) the underlying form
  $u$.  In general, we expect that each candidate $x \in \Gen(u)$
  encodes an alignment of the underlying form $u$ with some possible
  surface form $s$, and $C_j(x)$ evaluates this {\em pair} on some
  criterion.}  Then $u$ is redundant and we may simply take the
candidate $x$ to be the grammatical form.  Now the language $L(\Gram)$
is simply the image of \Lex under \Gram.  We will write $u_x$ for the
underlying form, if any, such that $x \in \Gen(u_x)$.

An {\bf attested form} of the language is a candidate $x$ that the
learner knows to be grammatical (i.e., $x \in L(\Gram)$).  $y$ is a
{\bf competitor} of $x$ if they are both in the same candidate set:
$u_x = u_y$.  If $x,y$ are competitors with $\vec{C}(y) < \vec{C}(x)$,
we say that $y$ {\bf beats} $x$ (and then $x$ is not optimal).

An ordinary learner does not have access to attested forms, since
observing that $x \in L(\Gram)$ would mean observing an utterance's
entire prosodic structure and underlying form, which ordinarily are
not vocalized.  An {\bf attested set} of the language is a set $X$
such that the learner knows that some $x \in X$ is grammatical (but
not necessarily {\em which} $x$).  The idea is that a set is attested
if it contains all possible candidates that are consistent with
something a learner heard.%
\footnote{This is of course a simplification.  Attested sets
corresponding to {\em laugh} and {\em laughed} can represent the
learner's uncertainty about the respective underlying forms, but not
the knowledge that the underlying forms are {\em related}.  In this
case, we can solve the problem by packaging the entire morphological
paradigm of {\em laugh} as a single candidate, whose attested set is
constrained by the two surface observations {\em and} by the
requirement of a shared underlying stem.  (A $k$-member paradigm may
be encoded in a form suitable to a finite-state system by interleaving
symbols from $2k$ aligned tapes that describe the $k$ underlying and
$k$ surface forms.)  Alas, this scheme only works within
disjoint finite paradigms: while it captures the shared underlying
stem of {\em laugh} and {\em laughed}, it ignores the shared
underlying {\em suffix} of {\em laughed} and {\em frowned}.%
}  
An {\bf attested surface set}---the case considered in this paper---is
an attested set all of whose elements are competitors; i.e., the
learner is sure of the underlying form but not the surface form.

Some computational treatments of OT place restrictions on the grammars
that will be considered.  The {\bf finite-state assumptions}
\cite{ellison-1994,eisner-1997-acl,frank-satta-1998,karttunen-1998,wareham-1998}
are that
\begin{itemize}
\negskip\item candidates and underlying forms are represented as strings
      over some alphabet;
\negskip\item \Gen is a regular relation;\footnote{Ellison \shortcite{ellison-1994} makes only
      the weaker assumption that $\Gen(u)$ is a regular set for each $u$.}
\negskip\item each $C_j$ can be implemented as a weighted
      deterministic finite-state automaton (WDFA) (i.e.,
      $C_j(x)$ is the total weight of the path accepting $x$ in the WDFA);
\negskip\item \Lex and any attested sets are regular.  
\end{itemize}
\negskip The {\bf bounded-violations assumption}
\cite{frank-satta-1998,karttunen-1998} is that the value of $C_j(x)$
cannot increase with $|x|$, but is bounded above by some $k$.

In this paper, we do not always impose these additional restrictions.
However, when demonstrating that problems are hard, we usually adopt
both restrictions to show that the problems are hard even for the
restricted case.

Throughout this paper, we follow \TS in supposing that the learner
already knows the correct {\em set} of constraints $\Con =
\{C_1,C_2,\ldots C_n\}$, but must learn their order $\vec{C} =
\angles{C_1, C_2, \ldots C_n}$, known as a {\bf ranking} of $\Con$.
The assumption follows from the OT philosophy that \Con is universal
across languages, and only the order of constraints differs.  The
algorithms for learning a ranking, however, are designed to be general
for any \Con, so they take \Con as an
input.\footnote{\label{footnote:structuredcon}That is, these methods
  are not tailored (as others might be) to exploit the structure of
  some specific, putatively universal \Con.  
  Hence they require time at least linear on $n=|\Con|$, if only to read
  all the constraints.  Given the variety of cross-linguistic
  constraints in the literature, one
  must worry: is $n$ huge?  Most authors
  following Ellison \shortcite{ellison-1994} allow as constraints all
  the regular languages over some alphabet $\Sigma$; then $n >
  s^{s(|\Sigma|-1)}$ distinct constraints can be described by DFAs of
  size $s$, where $\Sigma$ (or $s$) must be large to accommodate all
  features and prosodic constituents.
  One solution: let each constraint constrain only a few symbols in
  $\Sigma$ (e.g., bound the number of non-default transitions per DFA).  Indeed,
  Eisner
  \shortcite{eisner-1997-acl,eisner-1997-lsa}  proposes that \Con
  is the union of two ``primitive'' constraint families.  If each
  primitive constraint may mention at most $t$ of $T$ autosegmental tiers, then
  $n=O(T^t)$, which is manageable for small $t$.}

\section{RCD as Topological Sort}

\TS investigate the problem of ranking a constraint set
\Con given a set of attested forms $x_1, \ldots x_m$ and corresponding
competitors $y_1, \ldots y_m$.  The problem is to determine a ranking
$\vec{C}$ such that for each $i$, $\vec{C}(x_i) \leq \vec{C}(y_i)$
lexicographically.  Otherwise $x_i$ would be ungrammatical, as
witnessed by $y_i$.  

In this section we give a concise presentation and analysis of \TS's
{\bf Recursive Constraint Demotion (RCD)} algorithm for this problem.
Our presentation exposes RCD's connection to topological sort, from
which we borrow a simple bookkeeping trick that speeds it up.

\subsection{Compiling into Boolean Formulas}\label{sec:compile}

The first half of the RCD algorithm extracts the relevant information
from the $\{x_i\}$ and $\{y_i\}$, producing what \TS call {\em
mark-data pairs}.  We use a variant notation.  For each constraint $C
\in \Con$, we construct a negation-free, conjunctive-normal form (CNF)
Boolean formula $\phi(C)$ whose literals are other constraints:
$$\phi(C) = \bigwedge_{i: C(x_i) > C(y_i)} \; \bigvee_{C': C'(x_i) <
C'(y_i)} C'$$

The interpretation of the literal $C'$ in $\phi(C)$ is that $C'$
outranks $C$.  It is not hard to see that a constraint ranking is a
valid solution iff it satisfies $\phi(C)$ for every $C$.  For example,
if $\phi(d) = (a \vee b \vee c) \wedge (b \vee e \vee f)$, this means
that $d$ must be outranked by either $a, b$ or $c$ (else $x_1$ is
ungrammatical) and also by either $b, e$ or $f$ (else $x_2$ is
ungrammatical).

How expensive is this compilation step?  Observe that the inner term
$\bigvee_{C': C'(x_i) < C'(y_i)} C'$ is independent of $C$, so it only
needs to be computed and stored once.  Call this term $D_i$.  We first
construct all $m$ of the disjunctive clauses $D_i$, requiring time and
storage $O(mn)$.  Then we construct each of the $n$ formulas $\phi(C)
= \bigwedge_{i: C(x_i) > C(y_i)} D_i$ as a list of pointers to up to
$m$ clauses, again taking time and storage $O(mn)$.

The computation time is $O(mn)$ for the
steps we have already considered, but we must add $O(mnE)$, where $E$
is the cost of precomputing each $C(x_i)$ or $C(y_i)$ and may depend
on properties of the constraints and input forms.

We write $M (= O(mn))$ for the exact storage cost of the formulas,
i.e., $M = \sum_i |D_i| + \sum_C |\phi(C)|$ where $|\phi(C)|$ counts
only the number of conjuncts.  

\subsection{Finding a Constraint Ranking}

The problem is now to find a constraint ranking that satisfies
$\phi(C)$ for every $C \in \Con$.  Consider the special case where
each $\phi(C)$ is a simple conjunction of literals---that is,
$(\forall i)|D_i|=1$.  This is precisely the problem of topologically
sorting a directed graph with $n$ vertices and $\sum_C |\phi(C)| =
M/2$ edges.  The vertex set is \Con, and $\phi(C)$ lists the parents
of vertex $C$, which must all be enumerated before $C$.

Topological sort has two well-known $O(M+n)$ algorithms
\cite{cormen-leiserson-rivest-1990}.  One is based on depth-first
search.  Here we will focus on the other, which is: Repeatedly find a
vertex with no parents, enumerate it, and remove it and its outgoing
edges from the graph.

The second half of \TS's RCD algorithm is simply the obvious
generalization of this topological sort method (to directed
hypergraphs, in fact, formally speaking).
We describe it as a function
$\Rcd(\Con,\phi)$ that returns a ranking $\vec{C}$:
\begin{enumerate}
\item If $\Con=\emptyset$, return $\angles{}$.  Otherwise:
\vspace{-1pt}   
\item Identify a $C_1 \in \Con$ such that $\phi(C_1)$
is empty. ($C_1$ is surface-true, or ``undominated.'')  
\item If there is no such constraint, then fail: no ranking can be consistent
with the data.  
\item Else, for each $C \in \Con$, destructively remove
from $\phi(C)$ any disjunctive clause $D_i$ that mentions $C_1$.
\item Now recursively compute and return $\vec{C} =
\angles{C_1,\Rcd(\Con-\{C_1\},\phi)}$.
\end{enumerate}

Correctness of $\Rcd(\Con,\phi)$ is straightforward, by
induction on $n = |\Con|$.  The base case $n=0$ is trivial.  For $n>0$:
$\phi(C_1)$ is empty and therefore satisfied.  $\phi(C)$ is also
satisfied for all other $C$: any clauses containing $C_1$ are
satisfied because $C_1$ outranks $C$, and any other clauses are
preserved in the recursive call and therefore satisfied by the
inductive hypothesis.

We must also show completeness of $\Rcd(\Con,\phi)$: if there exists
at least one correct answer $\vec{B}$, then the function must not
fail.  Again we use induction on $n$.  The base case $n=0$ is trivial.
For $n>0$: Observe that $\phi(B_1)$ is satisfied in $\vec{B}$, by
correctness of $\vec{B}$.  Since $B_1$ is not outranked by anything,
this implies that $\phi(B_1)$ is empty, so \Rcd has at least one
choice for $C_1$ and does not fail.  It is easy to see that $\vec{B}$
with $C_1$ removed would be a correct answer for the recursive call,
so the inductive hypothesis guarantees that that call does
not fail either.

\subsection{More Efficient Bookkeeping}

\TS (p. 61) analyze the \Rcd function as taking time $O(mn^2)$;
in fact their analysis shows more precisely $O(Mn)$.
We now point out that careful bookkeeping can make it operate in time
$O(M+n)$, which is at worst $O(mn)$ provided $n>0$.  This means that
the whole RCD algorithm can be implemented in time $O(mnE)$, i.e., it
is bounded by the cost of applying all the constraints to all the
forms.

First consider the special case discussed above, topological sort.  In
linear-time topological sort, each vertex maintains a list of its
children and a count of its parents, and the program maintains a list
of vertices whose parent count has become 0.  The algorithm then
requires only $O(1)$ time to find and remove each vertex, and $O(1)$
time to remove each edge, for a total time of $O(M+n)$ plus $O(M+n)$
for initialization.

We can organize RCD similarly.  We change our representations (not
affecting the compilation time in \S\ref{sec:compile}).  Constraint
$C$ need not store $\phi(C)$.  Rather, $C$ should maintain a list of
pointers to clauses $D_i$ in which it appears as a disjunct (cf.\ ``a
list of its children'') as well as the integer $|\phi(C)|$ (cf.\ ``a
count of its parents'').  The program should maintain a list of
``undominated'' constraints for which $|\phi(C)|$ has become 0.
Finally, each clause $D_i$ should maintain a list of constraints $C$
such that $D_i$ appears in $\phi(C)$.

Step 2 of the algorithm is now trivial: remove the head $C_1$ of the
list of undominated constraints.  For step 4, iterate over the stored
list of clauses $D_i$ that mention $C_1$.  Eliminate each such $D_i$
as follows: iterate over the stored list of constraints
$C$ whose $\phi(C)$ includes $D_i$ (and then reset that list to
empty), and for each such $C$, decrement $|\phi(C)|$, adding $C$ to
the undominated list if $|\phi(C)|$ becomes 0.

The storage cost is still $O(M+n)$.  In particular, $\phi(C)$ is now
implicitly stored as $|\phi(C)|$ backpointers from its clauses $D_i$,
and $D_i$ is now implicitly stored as $|D_i|$ backpointers from its
disjuncts (e.g., $C_1$).  Since \Rcd removes each constraint
and considers each backpointer exactly once, in $O(1)$ time,
its runtime is $O(M+n)$.

In short, this simple bookkeeping trick eliminates RCD's quadratic
dependence on $n$, the number of constraints to rank.  As already
mentioned, the total runtime is now dominated by $O(mnE)$, the
preprocessing cost of applying all the constraints to all the input
forms.  Under the finite-state assumption, this can be be more tightly
bounded as $O(n\cdot\mbox{total size of input forms}) = O(n\cdot\sum_i
|x_i|+|y_i|)$, since the cost of running a form through a WDFA is
proportional to the former's length.

\subsection{Alternative Algorithms}\label{sec:cd}

\TS also propose an alternative to RCD called Constraint Demotion
(CD), which is perhaps better-known.  (They focus primarily on it, and
Kager's textbook \shortcite{kager-1999} devotes a chapter to it.)  A
disjunctive clause $D_i$ (compiled as in \S\ref{sec:compile}) is {\bf
processed} roughly as follows: for each $C$ such that $D_i$ is an
unsatisfied clause of $\phi(C)$, greedily satisfy it by demoting $C$
as little as possible.  CD repeatedly processes $D_1, \ldots D_m$
until all clauses in all formulas are satisfied.

CD can be efficiently implemented so that each pass through all
clauses takes time proportional to $M$.
But it is easy to construct datasets that require $n+1$ passes.  So the
ranking step can take time $\Omega(Mn)$, which contrasts unfavorably
with the $O(M+n)$ time for \Rcd.

CD does have the nice property (unlike RCD) that it maintains a
constraint ranking at all times.  An ``online'' (memoryless) version
of CD is simply to generate, process, and discard each clause $D_i$
upon arrival of the new data pair $x_i,y_i$; this converges, given
sufficient data.  But suppose one wishes to maintain a ranking that is
consistent with {\em all} data seen so far.  In this case, CD is
slower than RCD.  Modifying a previously correct ranking to remain
correct given the new clause $D_i$ requires at least one pass
through all clauses $D_1,\ldots D_i$ (as slow as RCD) and
up to $n+1$ passes (as slow as running CD on all clauses from
scratch, ignoring the previous ranking).

\section{Considering {\em All} Competitors}\label{sec:allcompet}

The algorithms of the previous section only ensure that each attested
form $x_i$ is at least as harmonic as a given competitor $y_i$:
$\vec{C}(x_i) \leq \vec{C}(y_i)$.  But for $x_i$ to be grammatical, it
must be at least as harmonic as {\em all} competitors.  We would like
a method that ensures this.  Such a method will rank a constraint set
\Con given only a set of attested forms $\{x_1,\ldots x_m\}$.

Like \TS, whose algorithm for this case is discussed in
\S\ref{sec:EDCD}, here we (dangerously) assume we have an efficient
computation of OT's production function $\Opt(\vec{C},u)$ (such as
Ellison \shortcite{ellison-1994}, Tesar \shortcite{tesar-1996}, or
Eisner \shortcite{eisner-1997-acl}).  This returns the subset of
$\Gen(u)$ on which $\vec{C}(\cdot)$ is lexicographically minimal,
i.e., the set of grammatical outputs for $u$.  For the analysis, let
$P$ be a bound on the runtime of our \Opt algorithm.  We will discuss
this runtime further in \S\ref{sec:gencomplex}!

\subsection{Generalizing RCD}\label{sec:rcdall}

We propose to solve this problem by running something 
like our earlier RCD algorithm, but considering all competitors at
once.

First, as a false start, let us try to construct the requirements
$\phi(C)$ in this case.  Consider the contribution of a single $x_i$
to a particular $\phi(C)$.  $x_i$ demands that for {\em any}
competitor $y$ such that $C(x_i) > C(y)$, $C$ must be outranked by
{\em some} $C'$ such that $C'(x_i) < C'(y)$.  One set of competitors
$y$ might all add the same clause $(a \vee b \vee c)$ to $\phi(C)$;
another set might add a different clause $(b \vee d \vee e)$.

The trouble here is that $\phi(C)$ may become intractably large.  This
will happen if the constraints are roughly orthogonal to one another.
For example, suppose the candidates are bit strings of length $n$, and
for each $k$, there exists a constraint $\Off_k$ preferring the $k$th
bit to be zero.\footnote{$\Off_k(x)$ simply extracts the $k$th bit of
$x$.  We will later denote it as $C_{\neg v_k}$.}  If $x_i =
1000\cdots 0$, then $\phi(\Off_1)$ contains all $2^{n-1}$ possible
clauses: for example, it contains $(\Off_2 \vee \Off_4 \vee \Off_5)$
by virtue of the competitor $y = 0101100000\cdots$.  Of course, the
conjunction of all these clauses can be drastically simplified in this
case, but not in general.

Therefore, we will skip the step of constructing formulas $\phi(C)$.
Rather, we will run something like \Rcd directly:
greedily select a constraint $C_1$ that does not eliminate any of the
attested forms $x_i$ (but that may eliminate some of its competitors),
similarly select $C_2$, etc.

In our new function $\RcdAll(\Con,\vec{B},\{x_i\})$, the input
includes a partial hierarchy $\vec{B}$ listing the constraints chosen
at previous steps in the recursion.  (On a non-recursive call,
$\vec{B}=\angles{}$.)  
\begin{enumerate}
\item If $\Con=\emptyset$, return $\angles{}$.  Otherwise:
\item By trying all constraints, find a constraint $C_1$ such that
      $(\forall i) x_i \in \Opt(\angles{\vec{B},C_1},u_{x_i})$
\item If there is no such constraint, then fail: no ranking can be consistent
      with the data.  
\item Else recursively compute and return $\vec{C} =
      \angles{C_1,\RcdAll(\Con-\{C_1\}, \angles{\vec{B},C_1},\{x_i\})}$
\end{enumerate}

It is easy to see by induction on $|\Con|$ that \RcdAll is
correct: if it does not fail, it always returns a ranking $\vec{C}$
such that each $x_i$ is grammatical under the ranking
$\angles{\vec{B},\vec{C}}$.  It is also complete, by the same argument
we used for \Rcd: if there exists a correct ranking, then there
is a choice of $C_1$ for this call and there exists a correct ranking
on the recursive call.

The time complexity of \RcdAll is $O(mn^2P)$.  Preprocessing and
compilation are no longer necessary (that work is handled by \Opt).
We note that if \Opt\ is implemented by successive winnowing of an
appropriately represented candidate set, as is common in finite-state
approaches, then it is desirable to cache the sets returned by \Opt
at each call, for use on the recursive call.  Then
$\Opt(\angles{\vec{B},C_1},u_{x_i})$ need not be computed from
scratch: it is simply the subset of $\Opt(\vec{B},u_{x_i})$ on which
$C_1(\cdot)$ is minimal.

\subsection{Alternative Algorithms}\label{sec:EDCD}

\TS provide a different, rather attractive solution to this problem,
which they call Error-Driven Constraint Demotion (EDCD).  This is
identical to the ``online'' CD algorithm of \S\ref{sec:cd}, except
that for each attested form $x$ that is presented to the learner, EDCD
automatically chooses a competitor $y \in \Opt(\vec{C},u_x)$,
where $\vec{C}$ is the ranking at the time.

If the supply of attested forms $x_1,\ldots x_m$ is limited, as
assumed in this paper, one may iterate over them repeatedly, modifying
$\vec{C}$, until they are all optimal.  When an attested form $x$ is
suboptimal, the algorithm takes time $O(nE)$ to compile $x,y$ into a
disjunctive clause and time $O(n)$ to process that clause using
CD.\footnote{Instead of using CD on the new clause only, one may use
RCD to find a ranking consistent with all clauses generated so far.
This step takes worst-case time $O(n^2)$ rather than $O(n)$ even with
our improved algorithm, but may allow faster convergence.
Tesar \shortcite{tesar-1997-ms} calls this version Multi-Recursive Constraint
Demotion (MRCD).}

\TS show that the learner converges after seeing at most $O(n^2)$
suboptimal attested forms, and hence after at most $O(n^2)$ passes
through $x_1,\ldots x_m$.  Hence the total time is $O(n^3E+mn^2P)$,
where $P$ is the time required by \Opt.  This is superficially worse
than our \RcdAll, which takes time $O(mn^2P)$, but really about as
good since $P$ dominates (see \S\ref{sec:gencomplex}).

Mainly, \RcdAll is simpler.
\S\ref{sec:rankcomplex} (note~\ref{fn:rcdallgood}) also shows that
\RcdAll needs less information from each call to \Opt; this improves
the complexity class of the call, though not of the full algorithm.

Algorithms that adjust constraint rankings or weights along a
continuous scale include the Gradual Learning Algorithm
\cite{boersma-1997}, which resembles simulated annealing, and maximum
likelihood estimation \cite{johnson-2000-jhu}.  These methods have the
considerable advantage that they can deal with noise and free variation in the
attested data.  Both algorithms repeat until convergence, which makes
it difficult to judge their efficiency except by experiment.

\section{Incompletely Observed Forms}\label{sec:attsetweak}

We now add a further wrinkle.  Suppose the input to the learner
specifies only $\Con$ together with attested surface sets
$\{X_i\}$, as defined in \S\ref{sec:formalism}, rather than attested forms.  This
version of the problem captures the learner's uncertainty about the
full description of the surface material.  As before, the goal is to
rank \Con in a manner consistent with the input.

With this wrinkle, even determining whether such a ranking exists
turns out to be surprisingly harder.  In \S\ref{sec:rankcomplex} we
will see that it is actually $\Sigma_2^p$-complete.  Here 
we only show it \NP-hard, using a construction that suggests
that the \NP-hardness stems from the need to consider exponentially
many rankings or surface forms.

\subsection{NP-Hardness Construction}\label{sec:unobshard}\label{sec:powerindex}

Given $r \in
\mathbb{N}$, we will be considering finite-state OT grammars of the
following form:
\begin{itemize}
\item $\Lex = \{\epsilon\}$.
\item $\Gen(\epsilon) = \Sigma^r$, the set of
all length-$r$ strings over the alphabet $\Sigma = \{1,2,\ldots r\}$.
(This set can be represented with a straight-line DFA of $r+1$ states
and $r^2$ arcs.) 
\item $\Con = \{\Early_j: 1 \leq j \leq r\}$, where for any $x \in
\Sigma^*$, 
the constraint $\Early_j(x)$ counts
the number of digits in $x$ before the first occurrence of digit $j$,
if any.  For example, $\Early_3(2188353) = \Early_3(2188) = 4$.  (Each
such constraint can be implemented by a WDFA of 2 states and $2r$ arcs.)
\end{itemize}
$\Early_j$ favors candidates in which $j$ appears early.
The ranking
$\langle\Early_5,\Early_8,$ $\Early_1,\ldots\rangle$ favors candidates
of the form $581\cdots$; no other candidate can be grammatical.

Given a directed graph $G$ with $r$ vertices identified by the digits
$1,2,\ldots r$.  A {\bf path} in $G$ is a string of digits $j_1 j_2
j_3\cdots j_k$ such that $G$ has edges from $j_1$ to $j_2$, $j_2$ to
$j_3$, \ldots\ and $j_{k-1}$ to $j_k$.  Such a string is called a {\bf
Hamilton path} if it contains each digit exactly once.  It is an
\NP-complete problem to determine whether an arbitrary graph $G$ has a
Hamilton path.

Suppose we let the attested surface set $X_1$ be the set of length-$r$
paths of $G$.  This is a regular set that can be represented in space
proportional to $r|G|$, by intersecting the DFA for $\Gen(\epsilon)$
with a DFA that accepts all paths of $G$.\footnote{The latter DFA is
  isomorphic to $G$ plus a start state.  The states are $0,1,\ldots
  r$;
  there is an arc from $j$ to $j'$ (labeled with $j'$) iff $j=0$
  or $G$ has an edge from $j$ to $j'$.}

Now $(\Con, \{X_1\})$ is an instance of the ranking problem whose size
is $O(r|G|)$.  We observe that any correct ranking algorithm
determines if $G$ has a Hamilton path.  Why?  A ranking is a vector
$\vec{C}=\angles{\Early_{j_1}, \ldots \Early_{j_r}}$, where
$j_1,\ldots j_r$ is a permutation of $1,\ldots r$.  The optimal form
under this ranking is in fact the string $j_1\cdots j_r$.  A string is
consistent with $X_1$ if it is a path of $G$, so the ranking $\vec{C}$
is consistent with $X_1$ iff $j_1\ldots j_r$ is a Hamilton path of
$G$.  If such a ranking exists, the algorithm is bound to find it, and
otherwise to return a failure code.  Hence the ranking problem of this
section is \NP-hard.

Further, if the Satisfiability Hypothesis (SH) holds
\cite{stearns-hunt-1990}, Hamilton Path must take time
$2^{\Omega(|G|)}$, {\em a fortiori} $2^{\Omega(r)}$.  Then any ranking
algorithm takes $2^{\Omega(n)}$ ($n=|\Con|$).

\subsection{Discussion}\label{sec:unobsharddiscuss}

Since each ranking of the constraints $\Early_j$ is trivial to test
against $X_1$ (by DFA intersection), the \NP-hardness of ranking them
arises not from the difficulty of each test (though other constraint
sets do have such hard tests! see \S\ref{sec:gencomplex}) but from the
$2^n$ possible rankings.  A brute-force check of exponentially many
rankings takes time $2^{\Theta(n)}$.
Thus, given SH, no ranking algorithm can
consistently beat such a brute-force check.

Note that our construction shows \NP-hardness for even a restricted
version of the ranking problem: finite-state grammars and finite
attested surface sets.  The result holds up even if we also make the
bounded-violations assumption (see \S\ref{sec:formalism}): the
violation count can stop at $r$, since $\Early_j$ need only work
correctly on strings of length $r$.  We revise the construction,
modifying the automaton for each $\Early_j$ by intersection (more or
less) with the straight-line automaton for $\Sigma^r$.  This preserves
$|\Con|$ and $X_1$ and blows up the ranker's input $\Con$ by only
$O(r)$.

By way of mitigating this stronger result, we note that the
construction in the previous paragraph bounds $|X_i|$ by $r!$ and the
number of violations by $r$.  These bounds (as well as $|\Con|=r$)
increase with the order $r$ of the input graph.  If the bounds were
imposed by universal grammar, the construction would not be possible
and \NP-hardness might not hold.  Unfortunately, any universal bounds
on $|X_i|$ or $|\Con|$ would hardly be small enough to protect the
ranking algorithm from having to solve huge instances of Hamilton
path.\footnote{We expect attested sets $X_i$ to be very
large---especially in the more general case where they reflect
uncertainty about the underlying form.  That is why we describe them
compactly by DFAs.  A universal constraint set $\Con$ would
also have to be very large 
(footnote~\ref{footnote:structuredcon}).}  As for bounded violations,
the only real reason for imposing this restriction is to ensure that
the OT grammar defines a regular relation
\cite{frank-satta-1998,karttunen-1998}.  In recent work, Eisner
\shortcite{eisner-2000-coling} argues that the restriction is too severe for
linguistic description, and proposes a more general class of
``directional constraints'' under which OT grammars remain 
regular.\footnote{Allowing directional constraints would not change
any of the classifications in this paper.}
If this relaxed restriction is substituted for a universal bound on
violations, the ranking problem remains \NP-hard, since each
$\Early_j$ is a directional constraint.

A more promising ``way out'' would be to universally restrict the size
or structure of the automaton that describes the attested set.  The
set used in our construction was quite artificial.  

However, in $\S\ref{sec:rankcomplex}$ we will answer all these
objections: we will show the problem to be $\Sigma_2^p$-complete,
using finite-state constraints with at most 1 violation (which,
however, will not interact as simply) and a natural attested set.

\subsection{Available Algorithms}

The \NP-hardness result above suggests that existing algorithms
designed for this ranking problem are either incorrect or intractable
on certain cases.  Again, this does not rule out efficient algorithms
for variants of the problem---e.g., for a specific universal 
$\Con$---nor does it rule out algorithms that
tend to perform well in the average case, or on small inputs, or on real
data.

\TS proposed an algorithm for this problem, RIP/CD, but left its
efficiency and correctness for future research (p. 39); Tesar and
Smolensky \shortcite{tesar-smolensky-2000} show that it is not
guaranteed to succeed.
Tesar \shortcite{tesar-1997-ms} gives a related algorithm based on
MRCD (see \S\ref{sec:EDCD}), but which sometimes requires iterating
over all the candidates in an attested surface set; this might easily
be intractable even when the set is finite.

\section{Complexity of OT Generation}\label{sec:gencomplex}

The ranking algorithms in \S\S\ref{sec:rcdall}--\ref{sec:EDCD} relied
on the existence of an algorithm to compute the independently
interesting ``language production'' function $\Opt(\vec{C},u)$, which
maps underlying $u$ to the set of optimal candidates in $\Gen(u)$.

In this section, we consider the computational complexity of some
functions related to \Opt:%
\footnote{All these functions take an additional argument \Gen, which we suppress for readability.}
\begin{itemize}
\negskip
\item $\OptVal(\vec{C},u)$: returns $\min_{x \in \Gen(u)} \vec{C}(x)$.
      This is the violation vector shared by all the optimal candidates $x \in
      \Opt(\vec{C},u)$.
\negskip
\item $\OptValZ(\vec{C},u)$: returns ``yes'' iff the last component
      of the vector $\OptVal(\vec{C},u)$ is zero.  This decision problem
      is interesting only because if it cannot be computed efficiently
      then neither can \OptVal (or \Opt).

\negskip
\item $\Beatable(\vec{C},u,\angles{k_1,\ldots k_n})$: returns ``yes'' iff
      $\OptVal(\vec{C},u) < \angles{k_1,\ldots k_n}$.
\negskip
\item $\Best(\vec{C},u,\angles{k_1,\ldots k_n})$: returns ``yes'' iff
      $\OptVal(\vec{C},u) = \angles{k_1,\ldots k_n}$.
\negskip
\item $\Check(\vec{C},x)$: returns ``yes'' iff $x \in
      \Opt(\vec{C},u_x)$.
      This checks whether an attested form is consistent with
      $\vec{C}$.    
\negskip
\item $\CheckSSet(\vec{C},X)$: returns ``yes'' iff $\Check(\vec{C},x)$
      for some $x\in X$.
      This checks whether an attested surface set (namely $X$)
      is consistent with $\vec{C}$.  
\end{itemize}
\negskip
These problems place a lower bound on the difficulty of OT generation, since an algorithm that found a
reasonable representation of $\Opt(\vec{C},u)$ (e.g., a DFA) could
solve them immediately, and an algorithm that found an exemplar $x \in
\Opt(\vec{C},u)$ could solve all but \CheckSSet immediately.  \S\ref{sec:rankcomplex}
will relate them to OT learning.

\subsection{Past Results}\label{sec:pastgenresults} 

Under finite-state assumptions, Ellison \shortcite{ellison-1994}
showed that for any fixed $\vec{C}$, a representation of
$\Opt(\vec{C},u)$ could be generated in time $O(|u| \log |u|)$, making
all the above problems tractable.  However, Eisner
\shortcite{eisner-1997-acl} showed generation to be intractable when
$\vec{C}$ was not fixed, but rather considered to be part of the
input---as when generation is called by an algorithm like \RcdAll that
learns rankings.  Specifically, Eisner showed that \OptValZ is
\NP-hard.  Similarly, Wareham \shortcite[theorem 4.6.4]{wareham-1998}
showed that a version of \Beatable is \NP-hard.\footnote{Wareham also
  gave hardness results for versions of \Beatable where some
  parameters are bounded or fixed.}  (We will obtain more precise
classifications below.)

To put this another way, the worst-case complexity of generation
problems is something like $O(|u| \log |u|)$ times a term exponential
in $|\vec{C}|$.  

Thus there are {\em some} grammars for which generation is very
difficult by any algorithm.  So when testing exponentially many
rankings (\S\ref{sec:attsetweak}), a learner may need to spend
exponential time testing an individual ranking.

We offer an intuition as to why generation can be so hard.  In
successive-winnowing algorithms like that of \cite{eisner-1997-acl},
the candidate set begins as a large simple set such as $\Sigma^*$, and
is filtered through successive constraints to end up (typically) as a
small simple set such as the singleton $\{x_1\}$.  Both these sets can
be represented and manipulated as small DFAs.  The trouble is that 
intermediate candidate sets may be complex and require
exponentially large DFAs to represent.  (Recall that the intersection
of DFAs can grow as the product of their sizes.)

For example, Eisner's \shortcite{eisner-1997-acl} \NP-hardness
construction led to such an intermediate candidate set, consisting of
all permutations of $r$ digits.  Such a set arises simply from a
hierarchy such as $\angles{\Project_1,\ldots\Project_r,\Short}$, where
$\Project_j(x) = 0$ provided that $j$ appears (at least once) in $x$,
and $\Short(x) = |x|$.  (Adding a bottom-ranked constraint that prefers $x$ to
encode a path in a graph $G$ forces \Opt\ to search for a
Hamilton path in $G$, which demonstrates \NP-hardness of \OptValZ.)  

\subsection{Relevant Complexity Classes}

Perhaps the reader recalls that $\Pol \subseteq \NP \cap \coNP
\subseteq \NP \cup \coNP \subseteq D^p \subseteq \Delta_2^p =
\Pol^{\sf NP} \subseteq \Sigma_2^p = \NP^{\sf NP}$.  If not, we will
review these classes as they arise.\footnote{Problems in all but \Pol
  are widely suspected to require exponential time---which suffices by
  brute-force search.  (Smaller classes allow ``more cleanly parallel''
  search.)}

These are classes of {\bf decision problems}, i.e., functions taking
values in \{yes,no\}.  Hardness and completness for such classes are
defined via many-one (Karp) reductions: $g$ is at least as hard as $f$
iff $(\forall x) f(x)=g(T(x))$ for some function $T(x)$ computable in
polynomial time.\footnote{$g$ is {\bf $X$-hard} if it is at least as hard as
all $f \in X$, and {\bf $X$-complete} if also $g\in X$.}

In contrast, \OptP is a class of integer-valued functions, introduced
by Krentel \shortcite{krentel-1988}.  Recall that \NP is the class of
decision problems solvable in polytime by a nondeterministic Turing
machine: each control branch of the machine checks a
different possibility and gives a yes/no answer, and the machine
returns the {\em disjunction} of the answers.  For \coNP, the machine
returns the {\em conjunction}.  For \OptP, each branch writes a binary
integer $\geq 0$, and the machine returns the {\em minimum} (or
maximum) of these answers.

A canonical example (analogous to \OptVal) is the Traveling
Salesperson problem---finding the minimum cost $\TspVal(G)$ of all
tours of an integer-weighted graph $G$.  It is \OptP-complete in the
sense that all functions $f$ in \OptP can be {\bf metrically reduced}
to it \cite[p.~493]{krentel-1988}.  A metric reduction solves an
instance of $f$ by transforming it to an instance of $g$ and then
transforming the integer result of $g$: $(\forall
x)f(x)=T_2(x,g(T_1(x)))$ for some polytime-computable functions $T_1:
\Sigma^* \rightarrow \Sigma^*$ and $T_2: \Sigma^* \times \mathbb{N}
\rightarrow \mathbb{N}$.

Krentel showed that $\OptP$-complete problems yield complete problems
for decision classes under broad conditions.  The question $\TspVal(G)
\leq k$ is of course the classical TSP decision problem, which is
\NP-complete.  (It is analogous to \Beatable.)  The reverse question
$\TspVal(G) \geq k$ (which is related to \Check) is \coNP-complete.
The question $\TspVal(G)=k$ (analogous to \Best) is therefore in the
class $D^p = \{L_1 \cap L_2: L_1 \in \NP$ and $L_2 \in \coNP\}$
\cite{papadimitriou-yannakakis-1982}, and it is complete for that
class.  Finally, suppose we wish to ask whether the optimal tour is
unique.  (Like \OptValZ and \CheckSSet, this asks about a complex
property of the optimum.)  Papadimitriou
\shortcite{papadimitriou-1984} first showed this question to be
complete for $\Delta_2^p = {\Pol}^{\sf NP}$, the class of languages
decidable in polytime by deterministic Turing machines that have
unlimited access to an oracle that can answer \NP questions in unit
time.  (Such a machine can certainly {\em decide} uniqueness: It can
compute the integer $\TspVal(G)$ by binary search, asking the oracle
for various $k$ whether or not $\TspVal(G) \leq k$, and then ask it a
final \NP question: do there exist two distinct tours with cost
$\TspVal(G)$?)

\subsection{New Complexity Results}\label{sec:gencomplexnew}

It is quite easy to show analogous results for OT generation.  Our
main tool will be one of Krentel's \shortcite{krentel-1988}
\OptP-complete problems: Minimum Satisfying
Assignment.  If $\phi$ is a CNF boolean formula on $n$ variables, then
$\Msa(\phi)$ returns the lexicographically minimal bitstring $b_1b_2
\cdots b_n$ that represents a satisfying assignment for $\phi$, or
$1^n$ if no such bitstring exists.\footnote{Krentel's presentation is actually in terms of
Maximum Satisfying Assignment, which merely reverses the roles of 0
and 1.  Also, Krentel does not mention that $\phi$ can be restricted
to CNF, but importantly for us, his proof of \OptP-hardness makes
this fact clear.} 

We consider only problems where we can compute $C_j(x)$, or determine
whether $x \in \Gen(u)$, in polytime.  We further assume that \Gen
produces only candidates of length polynomial in the size of the problem
input---or more weakly, that our functions need not produce correct
answers unless at least one {\em optimal} candidate is so bounded.

Our hardness results (except as noted) apply even to OT grammars with
the finite-state and bounded-violations assumptions
(\S\ref{sec:formalism}).  In fact, we will assume without further loss
of generality \cite{ellison-1994,frank-satta-1998,karttunen-1998} that
constraints are $\{0,1\}$-valued, hence representable by unweighted DFAs.

Notation: We may assume that all formulas $\phi$ use variables from a
set $\{v_1,v_2,\ldots v_{O(|\phi|)}\}$.  Let $\ell(\phi)$ be the
maximum $i$ such that $v_i$ appears in $\phi$.  We define the
constraint $C_\phi$ to map strings of at least $\ell(\phi)$ bits to
$\{0,1\}$, defining $C_\phi(b_1b_2\cdots)=0$ iff $\phi$ is true when
the variables $v_i$ in $\phi$ are instantiated respectively to values
$b_i$.  

If we do not make the finite-state assumptions, then any $C_\phi$ can
be represented trivially in size $|\phi|$.  But under these
assumptions, we must represent $C_\phi$ as a DFA that accepts just
those bitstrings that satisfy $\phi$.  While this is always
possible (operators $\wedge,\vee,\neg$ in $\phi$ correspond to DFA
operations), we necessarily take care in this case to use only
$C_\phi$ whose DFAs are polynomial in $|\phi|$.  In particular, if
$\phi$ is a disjunction of (possibly negated) literals, such as
$b_2\vee b_3\vee\neg b_7$, then a DFA of $\ell(\phi)+2$ states
suffices.

We begin by showing that $\OptVal(\vec{C},u)$ is \OptP-complete.  It
is obvious under our restrictions that it is in the class
\OptP---indeed it is a perfect example.  Each nondeterministic branch
of the machine considers some string $x$ of length $\leq p(|u|)$,
simply writing the bitstring $\vec{C}(x)$ if $x \in \Gen(u)$ and $1^n$
otherwise.

To show \OptP-hardness, we metrically reduce $\Msa(\phi)$ to \OptVal,
where $\phi=\bigwedge_{i=1}^m D_i$ is in CNF.  Let $r=\ell(\phi)$, and
put $\Lex = \{\epsilon\}$ and $\Gen(\epsilon) = \{0,1\}^r$.  Also put
$D_i' = D_i \vee (v_1 \wedge \ldots \wedge v_r)$, so that $1^r$ satisfies
each $C_{D_i'}$.  Now let $\vec{C}=\angles{C_{D_1'},\ldots$ $C_{D_m'},$
  $C_{\neg v_1},\ldots$ $C_{\neg v_r}}$.  Then 
$\Msa(\phi) =$ the last $r$ bits of 
$\OptVal(\vec{C}, \epsilon)$.\footnote{$C_{D_i'}$ requires a 
DFA of $2r+2$ states.  Remark: Without the finite-state
assumptions, we could just write $\Msa(\phi) =
\OptVal(\angles{C_{\phi\wedge\neg v_1},\ldots C_{\phi\wedge\neg
    v_r}},\epsilon)$ for any $\phi$.}

Because \OptVal is \OptP-complete, Krentel's theorem
3.1 says it is complete for $\FP^{\sf NP}$, the set of functions computable
in polynomial time using an oracle for \NP.  This is the function
class corresponding to the decision class $\Pol^{\sf NP} = \Delta_2^p$.

Next we show that $\Beatable(\vec{C},u,\vec{k})$ is \NP-complete.  It
is obviously in \NP.  To show \NP-hardness (and power index 1, so that
SH (\S\ref{sec:powerindex}) implies runtime $2^{\Omega({\rm size\ of\ 
    input})}$), again put $\phi=\bigwedge_{i=1}^m D_i$,
$r=\ell(\phi)$, and $\Gen(\epsilon)=\{0,1\}^r$.  Now $\CNFSat(\phi) =
\Beatable(\angles{C_{D_1},\ldots C_{D_m}},\epsilon,\angles{0,0,\ldots
  0,1})$.

Next consider $\Check(\vec{C},x)$.  This is simply
$\neg\Beatable(\vec{C},u_x,\vec{C}(x))$.  Even when restricted to
calls of this form, \Beatable remains just as hard.  To show this, we
tweak the above construction so we can write $\vec{C}(x)$ (for some
$x$) in place of $\angles{0,0,\ldots 0,1}$.  Add the new element
$\epsilon$ to $\Gen(\epsilon)$, and extend the constraint definitions
by putting $C_{D_i}(\epsilon)=0$ iff $i<m$.  Then $\CNFSat(\phi) =
\Beatable(\vec{C},\epsilon,\vec{C}(\epsilon))$.  Therefore $\Check=\neg\Beatable$ is
\coNP-complete.

Next we consider $\Best(\vec{C},u,\vec{k})$.  This problem is in $D^p$
for the same simple reason that the question $\TspVal(G)=k$ is (see
above).  If we do not make the finite-state assumptions, it is also
$D^p$-hard by reduction from the $D^p$-complete language $\SatUnsat =
\{(\phi,\psi): \phi\in\Sat, \psi\not\in\Sat\}$
\cite{papadimitriou-yannakakis-1982}, as follows:
$\SatUnsat(\phi,\psi) =
\Best(\angles{C_\phi,C_\psi},\epsilon,\angles{0,1})$, renaming variables as necessary so that $\phi$ uses only $v_1,\ldots
v_r$ and $\psi$ uses only $v_{r+1},\ldots v_{s}$, and
$\Gen(\epsilon)=\{0,1\}^{r+s}$.  

It is not clear whether \Best remains $D^p$-hard under the finite-state
assumptions.  But consider a more flexible variant
$\Range(\vec{C},u,\vec{k_1},\vec{k_2})$ that asks whether
$\OptVal(\vec{C},u)$ is between $\vec{k_1}$ and $\vec{k_2}$ inclusive.
This is also in $D^p$, and is $D^p$-hard because
$\SatUnsat(\phi\#\psi) = $ $\Range(\angles{C_{D_1},\ldots C_{D_m},
C_{D'_1},\ldots C_{D'_{m'}}}$, $\epsilon$, $\langle 0,\ldots$ $0,
0,\ldots 1\rangle, \angles{0,\ldots 0, 1,\ldots 1}$, where $\phi$, $\psi$, \Gen are as before and $\phi=\bigwedge_{i=1}^m D_i$,
$\psi=\bigwedge_{i=1}^{m'} D'_i$.

Finally, we show that the decision problems $\CheckSSet$ and
$\OptValZ$ are $\Delta_2^p$-complete.  They are in $\Delta_2^p$ by an
algorithm similar to the one used for TSP uniqueness above: since
\Beatable can be determined by an \NP oracle, we can find
$\OptVal(\vec{C},u)$ by binary search.\footnote{This takes
  polynomially many steps {\em provided} that $\log C_i(x)$ is
  polynomial in $|x|$ (as it is under the finite-state assumptions).
  We've already assumed that $|x|$ itself is polynomial on the input
  size, at least for optimal $x$.}  An additional call to an \NP
oracle decides $\CheckSSet(\vec{C},X)$ by asking whether $\exists x\in
X$ such that $\vec{C}(x)=\OptVal(\vec{C},u)$.  Such a call also
trivially decides $\OptValZ$.

The reduction to show $\Delta_2^p$-hardness is from a
$\Delta_2^p$-complete problem exhibited by Krentel \shortcite[theorem
3.4]{krentel-1988}: \Msalsb accepts $\phi$ iff the final (least
significant) bit of $\Msa(\phi)$ is 0.  Given $\phi$, we use the same
grammar as when we reduced \Msa to \OptVal: since \Msa and \OptVal
then share the same last bit,
$\Msalsb(\phi) = \OptValZ(\vec{C},\epsilon) = \CheckSSet(\vec{C}, \{0,1\}^{m+r-1}0)$.

Note that we did not have to use an unnatural attested surface set
as in \S\ref{sec:unobshard}.  The set $\{0,1\}^{m+r-1}0$ means that
the learner has observed only certain bits of the utterance---exactly
the kind of partial observation that we expect.  So even some
restriction to ``reasonable'' attested sets is unlikely to help.

\negskip\section{Complexity of OT Ranking}\label{sec:rankcomplex}

We now consider two ranking problems.  These ask whether \Con can
be ranked in a manner consistent with attested forms or attested sets:
\begin{itemize}
\negskip\item $\Rankable(\Con,\{x_1,\ldots x_m\})$: returns ``yes'' iff there is a
ranking $\vec{C}$ of \Con such that $\Check(\vec{C},x_i)$ for all $i$.
\item $\RankableSSet(\Con,\{X_i,\ldots X_m\})$: returns ``yes'' iff there is a
ranking $\vec{C}$ of \Con such that $\CheckSSet(\vec{C},X_i)$ for all $i$.
\end{itemize}

\negskip We do not have an exact classification of \Rankable at this
time.  But interestingly, the special case where $m=1$ and the
constraints take values in $\{0,1\}$ (which has sufficed to show most
of our hardness results) is only \coNP-complete---the same as \Check,
which merely verifies a solution.  Why?  Here \Rankable need only ask
whether there exists any $y \in \Gen(u_{x_1})$ that satisfies a proper
superset of the constraints that $x_1$ satisfies.  For if so, $x_1$
cannot be optimal under any ranking, and if not, then we can simply
rank the constraints that $x_1$ satisfies above the others.  This
immediately implies that the special case is in \coNP.  It also
implies it is \coNP-hard: using the grammar from our proof that \Check
is \coNP-hard (\S\ref{sec:gencomplexnew}), we write
$\CNFSat(\phi)=\neg\Rankable(\Con,\{\epsilon\})$.

The \RcdAll algorithm of \S\ref{sec:allcompet} provides an upper bound
on the complexity of \Rankable.  We saw in \S\ref{sec:rcdall} that
\RcdAll can decide \Rankable with $O(n^2m)$ calls to \Opt (where
$n=|\Con|$).  In fact, it suffices to call \Check rather than \Opt
(since \RcdAll only tests whether $x_i \in \Opt(\cdots)$).  Since
$\Check \in \coNP$, it follows that \Rankable is in ${\Pol}^{\sf coNP}
= {\Pol}^{\sf NP} = \Delta_2^p$.\footnote{\label{fn:rcdallgood}Tesar's
  EDCD and MRCD algorithms (\S\ref{sec:EDCD}) also run in polytime
  given an \NP oracle.  They too decide \Rankable with polynomially
  many calls to \Opt.  While they cannot substitute \Check for \Opt,
  they can substitute \OptVal (since they need optimal $y$ only to
  compute $\vec{C}(y)$).  Each call to $\OptVal \in {\FP}^{\sf NP}$
  can then be replaced by polynomially many calls to $\Check \in
  \coNP$.
  
  It is not relevant to \RcdAll vs.\@ EDCD that calling $\Check$ {\em
    once} (\coNP-complete) is in an easier complexity class than
  calling $\OptVal$ {\em once} (${\FP}^{\sf NP}$-complete).  Nor is it
  relevant for any practical purpose, since these two classes collapse
  under Turing (Cook) reductions.}
  
\RankableSSet is certainly in $\Sigma_2^p$, since it may be phrased in
$\exists\forall$ form as
$(\exists \vec{C}, \{x_i \in X_i\})$ $(\forall i, y_i \in
\Gen(u_{x_i}))$ $\vec{C}(x_i) \leq \vec{C}(y_i)$.  We saw in
\S\ref{sec:attsetweak} that it is \NP-hard even when the constraints
interact simply.  One suspects it is $\Delta_2^p$-hard, since merely
verifying a solution (i.e., \CheckSSet) is $\Delta_2^p$-complete
(\S\ref{sec:gencomplexnew}).  We now show that is actually
$\Sigma_2^p$-hard and therefore $\Sigma_2^p$-complete.

The proof is by reduction from the canonical $\Sigma_2^p$-complete
problem $\QSatTwo(\phi,r)$, where $\phi=\bigwedge_{i=1}^m
D_i$ is a CNF formula with $\ell(\phi) \geq r \geq 0$.  This returns
``yes'' iff \negskip
$$\negskip\exists b_1,\ldots b_r \neg\exists b_{r+1},\ldots b_s
\phi(b_1,\ldots b_s),$$ 
where $s \defeq \ell(\phi)$ and
$\phi(b_1,\ldots b_s)$ denotes the truth value of $\phi$ when the
variables $v_1,\ldots v_s$ are bound to the respective binary values
$b_1\ldots b_s$.

Given an instance of $\QSatTwo$ as above, put $\Lex = \{\epsilon\}$
and $\Gen(\epsilon) = \{0,1\}^{r+s} \cup X$ where $X=$ the set
$\{0,1\}^r2$.  Let $\Con = \{C_{D_1},\ldots C_{D_m}$, $C_{v_1},\ldots
C_{v_r}$, $C_{\neg v_1},\ldots C_{\neg v_r},\bar{X}\}$, where all
constraints have range $\{0,1\}$, we extend $C_{D_i}$ over $X$ by
defining it to be satisfied (i.e., take value 0) on all candidates in
$X$, and we define $\bar{X}$ to be satisfied on exactly those
candidates not in $X$.  As before, $C_{v_i}$ and $C_{\neg v_i}$ are
satisfied on a candidate iff its $i^{\rm th}$ bit is 1 or 0
respectively, regardless of whether the candidate is in $X$.

We now claim that $\QSatTwo(\phi,r) = \RankableSSet(\Con, \{X\})$.
The following terminology will be useful in proving this: Given a bit
sequence $\vec{b}=b_1,\ldots b_r$, define a {\bf $\vec{b}$-satisfier}
to be a bit string $b_1\cdots b_r b_{r+1}\ldots b_s$ such that
$\phi(b_1,\ldots b_s)$.  For $1 \leq i \leq r$, let $B_i, \bar{B}_i$
denote the constraints $C_{v_i}, C_{\neg v_i}$ respectively if
$b_i=1$, or vice-versa if $b_i=0$.  We then say that a ranking
$\vec{C}$ of \Con is {\bf $\vec{b}$-compatible} if $B_i$ precedes
$\bar{B}_i$ in $\vec{C}$ for every $1 \leq i \leq r$.

First observe that a candidate $y \in \Gen(\epsilon)$ is a
$\vec{b}$-satisfier iff it satisfies the constraints $B_1,\ldots B_r$
and $C_{D_1},\ldots C_{D_m}$ and $\bar{X}$.  From this it is not
difficult to see that if $\vec{C}$ is a $\vec{b}$-compatible ranking,
then $y$ beats $x$ (i.e., $\vec{C}(y) < \vec{C}(x)$) for any
$\vec{b}$-satisfier $y$ and any $x \in X$.\footnote{$y$ satisfies
  $\bar{X}$ while $x$ doesn't, so $\vec{C}(y)\neq
  \vec{C}(x)$.  And $\vec{C}(y)
  > \vec{C}(x)$ is impossible, for if $x$ satisfies any constraint
  that $y$ violates, namely some $\bar{B}_i$, then it violates a
  higher-ranked constraint that $y$ satisfies, namely $B_i$.}
Now for the proof:

Suppose $\RankableSSet(\Con,$ $\{X\})$.  Then choose $x \in X$ and
$\vec{C}$ a ranking of \Con such that $x$ is optimal (i.e.,
$\Check(\vec{C},x)$).  For each $1 \leq i \leq r$, let $b_i=1$ if
$C_{v_i}$ is ranked before $C_{\neg v_i}$ in $\vec{C}$, otherwise
$b_i=0$.  Then $\vec{C}$ is a $\vec{b}$-compatible ranking.  Since $x
\in X$ is optimal, there must be no $\vec{b}$-satisfiers $y$, i.e.,
$\QSatTwo(\phi,r)$.

Conversely, suppose $\QSatTwo(\phi,r)$.  This means we can choose
$b_1,\ldots b_r$ such that there are no $\vec{b}$-satisfiers.  Let
$\vec{C}=\angles{C_{D_1},\ldots C_{D_m}$, $B_1,\ldots B_r$,
$\bar{B}_1,\ldots \bar{B}_r, \bar{X}}$.  Observe that $x =
b_1\cdots b_r2 \in X$ satisfies the first $m+r$ of the constraints;
this is optimal (i.e., $\Check(\vec{C},x)$), since any better
candidate would have to be a $\vec{b}$-satisfier.\footnote{Since it
would have to satisfy the first $m+r$ constraints plus a later
constraint, which could only be $\bar{X}$.}  Hence there is a ranking
$\vec{C}$ consistent with $X$, i.e., 
$\RankableSSet(\Con,\{X\})$.

\negskip\section{Optimization vs. Derivation}

The above results mean that OT generation and ranking are hard.  We
will now see that they are harder than the corresponding problems in
deterministic derivational theories, assuming that the complexity
classes discussed are distinct.

A {\bf derivational grammar} consists of the following elements (cf. \S\ref{sec:formalism}):
\begin{itemize}
\negskip\item an alphabet $\Sigma$;
\negskip\item a set $\Lex\subseteq \Sigma^*$ of underlying forms;
\negskip\item a vector $\vec{R} = \angles{R_1,\ldots R_n}$ of {\bf rules},
      each of which is a function from $\Sigma^*$ to $\Sigma^*$.
\end{itemize}

\negskip The grammar maps each $x \in \Lex$ to $\vec{R}(x) \defeq R_n
\circ \cdots \circ R_2 \circ R_1(x)$.  If all the rules are
polytime-computable (i.e., in the function class \FP), then so is
$\vec{R}$.  (By contrast, the OT analogue \Opt is complete for the
function class $\FP^{\sf NP}$.)  It follows that the derivational
analogues of the decision problems given at the start of
\S\ref{sec:gencomplex} are in \mbox{\sf P}\footnote{However, Wareham
\shortcite{wareham-1998} analyzes a more powerful derivational
approach where the rules are nondeterministic: each $R_i$ is a
relation rather than a function.  Wareham shows that generation in
this case is \NP-hard (Theorem 4.3.3.1).  He does not consider
learning.}  (whereas we have seen that the OT versions range from
\NP-complete to $\Delta_2^p$-complete).

How about learning?  The {\bf rule ordering problem} $\OrderableSSet$
takes as input a set $\Rul$ of possible rules, a unary integer $n$,
and a set of pairs $\{(u_1,X_1),\ldots (u_m,X_m)\}$ where
$u_i \in \Sigma^*$ and $X_i \subseteq \Sigma^*$.  It returns ``yes''
iff there is a a rule sequence $\vec{R} \in \Rul^n$ such that
$(\forall i)\vec{R}(u_i) \in X_i$.  It is clear that this problem is
in \NP.  This makes it easier than its OT analogue $\RankableSSet$ and
possibly easier than $\Rankable$.

For interest, we show that \OrderableSSet is \NP-complete, as is its
restricted version \Orderable (where the attested sets $X_i$ are
replaced by attested forms $x_i$).  As usual, our result holds even
with finite-state restrictions: we can require the rules in \Rul to be
regular relations \cite{johnson-1972}.  The hardness proof is by
reduction from Hamilton Path (defined in \S\ref{sec:unobshard}).
Given a directed graph $G$ with vertices $1,2,\ldots n$, put
$\Sigma=\{\#,0,1,2,\ldots n\}$.  Each string we consider will be
either $\epsilon$ or a permutation of $\Sigma$.  Define $\Move_j$ to
be a rule that maps $\alpha j \beta \# \gamma i$ to $\alpha \beta \#
\gamma i j$ for any $i,j \in \Sigma$, $\alpha,\beta,\gamma \in
\Sigma^*$ such that $i=0$ or else $G$ has an edge from $i$ to $j$, and
acts as the identity function on other strings.  Also define $\Accept$
to be a rule that maps $\# \alpha$ to $\epsilon$ for any $\alpha \in
\Sigma^*$, and acts as the identity function on other strings.  Now
$\Orderable(\{\Move_1,\ldots \Move_n,\Accept\},n+1,\{(12\cdots
n\#0,\epsilon)\})$ decides whether $G$ has a Hamilton path.

\section{Conclusions}

See the abstract for our most important results.  Our main conclusion
is a warning that OT carries large computational burdens.  When
formulating the OT learning problem, even small nods in the direction
of realism quickly drive the complexity from linear-time up through
\coNP (for multiple competitors) into the higher complexity classes
(for multiple possible surface forms).

Hence all OT generation and learning algorithms should be suspect.
Either they oversimplify their problem, or they sometimes fail, or
they take worse than polynomial time on some class of inputs.  (Or
they demonstrate $\Pol=\NP$!)

One constraint ranking problem we consider, \RankableSSet, is in fact
a rare ``natural'' example of a problem that is complete for the
higher complexity class $\Sigma_2^p$ (``$\exists\forall$'').
Intuitively, an OT learner must both pick a constraint ranking
($\exists$) and check that an attested form beats or ties all
competitors under that ranking ($\forall$).
Some other learning problems were already known to be
$\Sigma_2^p$-complete \cite{ko-tzeng-1991}, but ours differs in that
the input has no negative exemplars (not even implicit ones, given ties).

This paper leaves some theoretical questions open.  Most important is
the exact classification of \Rankable.  Second, we are interested in
any cases where problem variants (e.g., accepting vs.\@ rejecting the
finite-state assumptions) differ in complexity.  Third, in the same
spirit, parameterized complexity
analyses \cite{wareham-1998} may help further identify sources of hardness.

We are also interested in more realistic versions of the
phonology learning problem.  We are especially interested in the
possibility that $\Con$ has internal structure, as discussed in
footnote \ref{footnote:structuredcon}, and in the problem of learning
from general attested sets, not just attested surface sets.

Finally, in light of our demonstrations that efficient algorithms are
highly unlikely for the problems we have considered, we ask: Are there
restrictions, reformulations, or randomized or approximate methods
that could provably make OT learning practical in some sense?

{\small
\bibliographystyle{acl}  
\bibliography{/u/jason/info/eisner}    
}

\end{document}